\documentclass[conference]{IEEEtran}
\IEEEoverridecommandlockouts
\usepackage{cite}
\usepackage{amsmath,amssymb,amsfonts}
\usepackage{algorithmic}
\usepackage{graphicx}
\usepackage{textcomp}
\usepackage{xcolor}
\usepackage{array}
\usepackage{booktabs}
\usepackage{multirow}
\usepackage{notes2bib}
\def\BibTeX{{\rm B\kern-.05em{\sc i\kern-.025em b}\kern-.08em
    T\kern-.1667em\lower.7ex\hbox{E}\kern-.125emX}}
\begin{document}

\title{Domain-Specific Language Model Post-Training for Indonesian Financial NLP}

\author{\IEEEauthorblockN{
Ni Putu Intan Maharani\textsuperscript{\textbf{1}},
Yoga Yustiawan\textsuperscript{\textbf{2}},
Fauzy Caesar Rochim\textsuperscript{\textbf{2}}, Ayu Purwarianti\textsuperscript{\textbf{1}}
}
\IEEEauthorblockA{
\textsuperscript{1}School of Electrical Engineering and Informatics \\
Institut Teknologi Bandung\\ 
Bandung, Indonesia \\
23522048@std.stei.itb.ac.id, ayu@itb.ac.id}
\IEEEauthorblockA{
\textsuperscript{2}BRIBRAIN Academy\\
PT Bank Rakyat Indonesia (Persero) Tbk\\
Jakarta, Indonesia \\
yoga.yustiawan@work.bri.co.id, fauzy.caesar@work.bri.co.id}
}

\maketitle

\begin{abstract}
BERT and IndoBERT have achieved impressive performance in several NLP tasks. There has been several investigation on its adaption in specialized domains especially for English language. We focus on financial domain and Indonesian language, where we perform post-training on pre-trained IndoBERT for financial domain using a small scale of Indonesian financial corpus. In this paper, we construct an Indonesian self-supervised financial corpus, Indonesian financial sentiment analysis dataset, Indonesian financial topic classification dataset, and release a family of BERT models for financial NLP. We also evaluate the effectiveness of domain-specific post-training on sentiment analysis and topic classification tasks. Our findings indicate that the post-training increases the effectiveness of a language model when it is fine-tuned to domain-specific downstream tasks.
\end{abstract}


\begin{IEEEkeywords}
domain-specific language model, post-trained language model, financial NLP, sentiment analysis, topic classification
\end{IEEEkeywords}

\section{Introduction}
Data is driving finance nowadays, and the most important data may be found in textual form, such as documents, texts, websites, forums, and other places\footnote{https://www.analyticssteps.com/blogs/6-applications-nlp-finance}. Moreover, banking institutions in Indonesia are increasingly employing textual data and NLP techniques to create a financial infrastructure that can make a more data-driven and intelligent decision. 

In natural language processing (NLP), pretraining large neural language models using unlabeled text on either one language or multiple languages has proven to be a successful method for transfer learning. One of the notable examples is Bidirectional Encoder Representations from Transformers (BERT), which has become a standard benchmark for training NLP models for various downstream tasks. Another example is IndoBERT, the implementation of BERT specific for Indonesian language which also performs well as a building block for training task-specific NLP models for Indonesian language \cite{wilie-etal-2020-indonlu}. However, those pre-training works focus on the general domain in which the unlabeled text data are collected from Web domains, newswire, Wikipedia, and BookCorpus \cite{wilie-etal-2020-indonlu, devlin-etal-2019-bert}. Previous studies have explored on a specific domain such as financial \cite{huang-finbert, yang2020finbert, araci2019finbert}, but their focus is on the implementation using English language.

To fill the gap, we perform continual pre-training or post-training on IndoBERT (BERT implementation for Indonesian) with an Indonesian self-supervised financial corpus, including financial news articles and corporate financial reports. We also evaluate its performance by applying the post-trained domain-specific models for sentiment analysis task which is a strongly domain-dependent task \cite{yang2020finbert} and financial topic classification task. This paper reports about our investigation into domain-specific post-training for financial domain in Indonesian language and the following contributions: (1) Indonesian self-supervised financial corpus, consisting of texts from financial news articles and corporate financial reports. (2) Indonesian financial sentiment analysis dataset, consisting of texts from news article titles that refer to a specific financial institution including its sentiment labels (IndoFinSent) and a translated version of Financial Phrasebank \cite{malo2013good}. (3) Indonesian financial topic classification dataset, consisting of texts from Twitter that are related to financial domain (i.e., a translated version of the Twitter Financial News dataset\footnote{https://huggingface.co/datasets/zeroshot/twitter-financial-news-topic}). (4) A systematic evaluation on using domain-specific post-trained models for sentiment analysis and topic classification tasks in the financial domain.

\section{Related Works}
Recently, self-supervised pre-training of contextual language models on large general domain corpora, such as ELMo \cite{peters-etal-2018-deep}, ULM-Fit \cite{howard2018universal}, XLNet \cite{yang2020xlnet}, GPT \cite{radford2018improving}, BERT \cite{devlin-etal-2019-bert}, and IndoBERT \cite{wilie-etal-2020-indonlu} have significantly improved performance on various natural language processing downstream tasks, including sentence classification, token classification, and question answering. IndoBERT, as the foundation of this research, is an implementation of BERT in Indonesian language. IndoBERT has similar model architecture as BERT in which it is a multi-layer bidirectional Transformer encoder \cite{devlin-etal-2019-bert}. However, IndoBERT is pre-trained by using Indo4B dataset that consists of around 4B
words, with around 250M sentences that covers both formal and colloquial Indonesian texts \cite{wilie-etal-2020-indonlu}.

Moreover, it also has been shown that pre-training a language model using domain-specific corpora can further improve the downstream task performance than fine-tuning the generic language model. Previous studies have explored on specific domains such as biomedical \cite{gu-biomedical, Lee_2019, peng-etal-2019-transfer}, scientific domain \cite{beltagy-etal-2019-scibert}, legal domain \cite{chalkidis-etal-2020-legal}, and financial \cite{huang-finbert, yang2020finbert, araci2019finbert}. They use domain-specific corpora to pre-train the language model (mainly BERT) and evaluate its effectiveness on various downstream tasks. There are two pre-training paradigms used on those previous studies which are: (1) pre-training from scratch using domain-specific corpora; and (2) continual pre-training (post-training) from pre-trained generic language model using domain-specific corpora. In this paper, we perform continual pre-training (post-training) from pre-trained IndoBERT using our constructed financial corpus in Indonesian.

\section{Indonesian Self-Supervised Financial-Domain Corpus}

\subsection{Corpus Construction}
Our corpus is primarily based on 875 financial news articles from CNBC and Bisnis.com and corporate financial reports from three largest financial institutions in Indonesia. For the financial news articles, we use BeautifulSoup to scrape relevant news article contents from CNBC especially 'Market' and 'My Money' categories and from Bisnis.com especially '\textit{Perbankan}', '\textit{Asuransi}, 'Multifinance', 'Personal Finance', '\textit{Moneter}', '\textit{Bisnis Syariah}', and 'Fintech'. For the corporate financial reports, we transforms PDF files provided in each corporate's investor relations site to text files using an online tool PDF2Go\footnote{https://www.pdf2go.com/pdf-to-text}. We tokenize the texts by sentence in this case.

To ensure a clean corpus, we remove irrelevant texts such as (inline) advertisements and links to related articles. We also remove multiple white spaces and special characters. A random selection was reviewed to ensure high quality.

\begin{table}
\centering
\caption{Indonesian Self-Supervised Financial-Domain Corpus Statistics}
\label{table:1}
\begin{tabular}{p{1.25cm} p{1cm} p{1cm} p{1cm} p{1.25cm}}
\hline
\textbf{Dataset} & \textbf{\#Words} & \textbf{Size (MB)} & \textbf{Style} & \textbf{Source}\\
\hline
Financial News Articles & 357,113 & 2.372 & Mixed & CNBC, Bisnis.com\\
\hline
Financial Corporate Reports & 290,473 & 2.48 & Formal & Corporate's Investor Relations Site\\
\hline
\end{tabular}
\end{table}

\subsection{Corpus Statistics}
The statistics of our financial domain corpus can be found in Table \ref{table:1}. As mentioned, the corpus is collected from online news platforms and corporate financial reports. As the pre-training of IndoBERT used 23.43 GB of text data \cite{wilie-etal-2020-indonlu}, in this work, we collect a small scale of domain-specific text data with total size of 4.85 MB and 647,586 words\footnote{https://huggingface.co/datasets/intanm/financial\_news\_id\_v1.0}\footnote{https://huggingface.co/datasets/intanm/indonesian\_financial\_statements}.

\section{Indonesian Financial Downstream Tasks Dataset}
\label{sec:fin-sa-dataset}
\subsection{Financial Sentiment Analysis}
\subsubsection{Corpus Construction}
For financial sentiment analysis task, labeled dataset is required for training and evaluating the resulted models. We use Financial Phrasebank \cite{malo2013good}, a financial sentiment analysis dataset with three sentiment classes, namely, negative (0), neutral (1), and positive (2). We then translate the dataset into Indonesian language by using Google Translate\footnote{https://huggingface.co/datasets/intanm/indonesian-financial-phrasebank}.

Moreover, we also construct a financial sentiment analysis dataset (IndoFinSent)\footnote{https://huggingface.co/datasets/intanm/IndoFinSent} by scraping financial news article titles that are referring to a financial institution in Indonesia from the following sources, namely, Liputan6, CNBC, Detik, Sindo, Jawapos, and Metrotv News. Each of the sentence (news title) was manually labeled by 2 annotators and cross-checked by these annotators. In sum, we collect and label 2,274 entries in our dataset. The annotation took about 38 hours.

\begin{table}
\centering
\caption{Indonesian Financial Downstream Task Dataset Statistics}
\label{table:2}
\begin{tabular}{p{2cm}cccc}
\hline
\textbf{Dataset} & \textbf{Split} & \textbf{Neg. (0)} & \textbf{Neut. (1)} & \textbf{Pos. (2)} \\
\hline
\multirow{3}{2cm}{Translated Financial Phrasebank (ID)} & Train & 224 & 993 & 412 \\
                             & Val & 23 & 122 & 37 \\
                             & Test & 56 & 276 & 121 \\
\hline
\textbf{Total} & - & 303 & 1391 & 570 \\
\hline
\multirow{3}{2cm}{IndoFinSent} & Train & 326 & 408 & 857 \\
                             & Val & 54 & 67 & 106 \\
                             & Test & 97 & 122 & 237 \\
\hline
\textbf{Total} & - & 477 & 597 & 1200\\
\hline
\end{tabular}
\end{table}

\subsubsection{Corpus Statistics}
The statistics of the dataset is shown in Table \ref{table:2}. For the translated version of Financial Phrasebank \cite{malo2013good}\footnote{https://huggingface.co/datasets/financial\_phrasebank}, we use the "all agree" subset consisting of 2,264 entries of sentence in financial domain with its respective sentiment labels. Most of the sentences have neutral labels (61.44\%), followed by positive labels of 25.18\% and negative labels of 13.38\%. For IndoFinSent, it is dominated by positive label entries of 52.77\%, followed by neutral and negative label entries of 26.25\% and 20.98\% respectively. Examples of the dataset contents can be seen on Table \ref{table:3}.

\begin{table*}
\centering
\caption{Indonesian Financial Downstream Task Dataset Examples}
\label{table:3}
\begin{tabular}{p{2cm} p{12.5cm} p{1cm}}
\hline
\textbf{Dataset} & \textbf{Sentence} & \textbf{Sent. Label}\\
\hline
\multirow{8}{2cm}{Translated Financial Phrasebank (ID)} & "Incap Contract Manufacturing Services Private Limited telah menandatangani perjanjian dengan enam pelanggan baru di India." & 2\\
                    & \textit{"Incap Contract Manufacturing Services Private Limited has signed agreements with six new customers in India."} &   \\
                    &   &   \\
                    & "Total durasi proyek adalah tiga tahun dan bernilai sekitar EUR 11,5 m." & 1\\
                    & \textit{"The total duration of the project is three years and is valued at around EUR 11.5m."} &   \\
                    &   &   \\
                    & "Sebuah survei yang dilakukan oleh Taloustutkimus untuk Sampo Life menunjukkan bahwa perusahaan sangat tidak siap untuk kehilangan karyawan kunci." & 0\\
                    & \textit{"A survey conducted by Taloututkimus for Sampo Life shows that the company is very unprepared to lose key employees."} &   \\
\hline
\multirow{8}{2cm}{IndoFinSent} & "Kredit UMKM BRI Tembus Rp 989,6 Triliun pada Kuartal I 2023" & 2\\
                    & \textit{"BRI MSME Credit Reaches IDR 989.6 Trillion in the First Quarter of 2023"} &  \\
                    &   &   \\
                    & "Profil Awan Nurmawan Nuh, Irjen Kemenkeu yang Diangkat Jadi Komisaris BRI" & 1\\
                    & \textit{"Profile of Awan Nurmawan Nuh, Inspector General of the Ministry of Finance who was Appointed as Commissioner of BRI"} &  \\
                    &   &   \\
                    & "VIDEO: Nasabah BRI Mengeluh Saldo Rekening Berkurang hingga Jutaan Rupiah" & 0\\
                    & \textit{"VIDEO: BRI Customers Complain about Decreased Account Balances of up to Millions of Rupiah"} &   \\
\hline
\multirow{8}{2cm}{Translated Twitter Financial News (ID)} & ''Netflix dan rekan-rekannya bersiap untuk 'kembali ke pertumbuhan,' kata para analis, memberikan satu saham kenaikan 120\%'' & Analyst Update\\
                    & \textit{''Netflix and its peers are set for a 'return to growth,' analysts say, giving one stock 120\% upside''} &  \\
                    &   &   \\
                    & ''Inflasi Inggris datang dengan panas, melonjak ke level tertinggi 40 tahun sebesar 9,4\% @lizzzburden memiliki detailnya'' & Macro\\
                    & \textit{''UK inflation comes in hot, surging to a 40 year high of 9.4\%  @lizzzburden has the details''} &  \\
\hline
\end{tabular}
\end{table*}

\subsection{Financial Topic Classification}

\begin{table}
\centering
\caption{Indonesian Financial Topic Classification Dataset}
\label{table:topic-clf-dataset}
\begin{tabular}{p{4cm} p{1cm} p{1cm} p{1cm}}
\hline
\textbf{Class Label} & \textbf{\#Train} & \textbf{\#Val.} & \textbf{\#Test}\\
\hline
\textit{Top-4}\\
\hline
Company \& Product News & 2,474 & 852 & 1,071\\
Stock Commentary & 1,493 & 528 & 625\\
Macro & 1,279 & 415 & 543\\
General News \& Opinion & 1,107 & 336 & 448\\
\hline
\textit{Bottom-4}\\
\hline
Analyst Update & 190 & 73 & 65\\
Currencies & 111 & 32 & 55\\
Gold \& Metals \& Materials & 49 & 13 & 20\\
IPO & 32 & 14 & 12\\
\hline
Total & 11,891 & 4,117 & 5,097
\end{tabular}
\end{table}
\subsubsection{Corpus Construction}
In addition to sentiment analysis task in financial domain, we also perform topic classification task in the domain. We use the Twitter Financial News dataset\footnote{https://huggingface.co/datasets/zeroshot/twitter-financial-news-topic} annotated with 20 topics, namely, Analyst Update, Fed \& Central Banks, Company \& Product News, Treasuries \& Corporate Debt, Dividend, Earnings, Energy \& Oil, Financials, Currencies, General News \& Opinion, Gold \& Metals \& Materials, IPO, Legal \& Regulation, M\&A \& Investments, Macro, Markets, Politics, Personnel Change, Stock Commentary, and Stock Movement. This dataset was published in English with the total of 21,105 documents. We then translate the dataset into Indonesian language by using Google Translate. We clean the dataset by removing links in each of the document. We also create the train (56.34\%), validation (19.51\%), and test (24.15\%) split for the translated dataset.
\subsubsection{Corpus Statistics}
As it can be seen from Table \ref{table:topic-clf-dataset}, the dataset is dominated with Company \& Product News topic with the total of 4,397 (20.83\%) entries out of the whole dataset, followed by Stock Commentary, Macro, and General News \& Opinion with the total of 2,646 (12.54\%), 2,237 (10.60\%), 1,891 (8.96\%) respectively. Other topics' presence in the dataset are ranging from 0.27\% to 5.82\%. Examples of the dataset contents can also be seen on Table \ref{table:3}.

\section{Methodology}
\begin{table*}
\centering
\caption{The details of IndoBERT models used for domain-specific post-training \cite{wilie-etal-2020-indonlu}}
\label{table:4}
\begin{tabular}{lllllllll}
\hline
\textbf{Model} & \textbf{\#Params} & \textbf{\#Layers} & \textbf{\#Heads} & \textbf{Emb. Size} & \textbf{Hidden Size} & \textbf{FFN Size} & \textbf{Language Type} & \textbf{Pre-train Emb. Type}\\
\hline
IndoBERT (base) & 124.5M & 12 & 12 & 768 & 768 & 3072 & Mono & Contextual \\
IndoBERT (large) & 335.2M & 24 & 16 & 1024 & 1024 & 4096 & Mono & Contextual \\
\hline
\end{tabular}
\end{table*}
In this paper, we use pre-trained IndoBERT models as the starting point of our experiment as shown in Table \ref{table:4}.

\subsection{Baselines}
As a baseline, we directly fine-tune generic IndoBERT for sentiment analysis and topic classification (single sentence classification) tasks. The models are fine-tuned for 2 epochs with a learning rate of 2e-5 and a weight decay of 0.01 using the translated Financial Phrasebank dataset on GPU T4. We also fine-tuned the models for different training data sizes to compare post-training effectiveness given smaller training data.

\begin{figure*}[htbp]
\centerline{\includegraphics[width=0.75\textwidth]{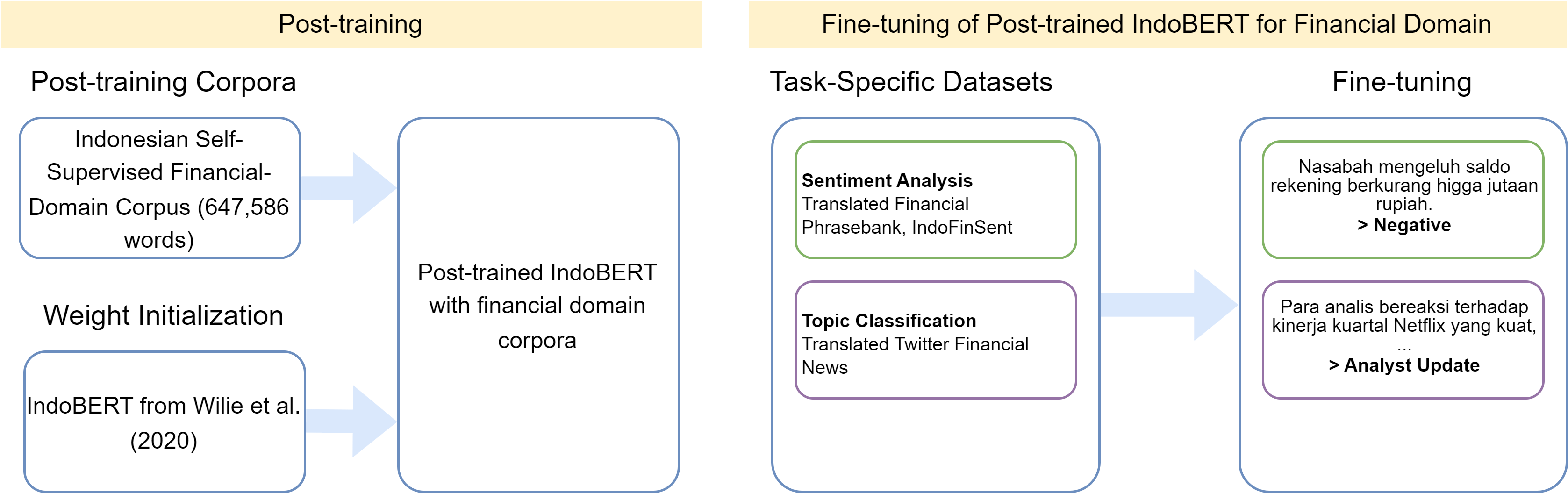}}
\caption{Overview of the domain-specific post-training and fine-tuning}
\label{fig-overview}
\end{figure*}

\subsection{Domain-Specific Post-Training}
The overview of domain-specific post-training and fine-tuning is shown in Figure \ref{fig-overview}. We perform post-training or continual pre-training from pre-trained IndoBERT (base and large architectures) by using the constructed self-supervised financial corpus. All models are post-trained on GPU T4. The training takes approximately 35 and 55 minutes on base and large architectures respectively. We use a batch size of 8 and a learning rate of 2e-5 for both architectures. Both base and large models are post-trained by using masked language modeling loss with 20 epochs and a weight decay of 0.01. For the large models, we first train it for 10 epochs and then continue for another 10 epochs due to limited resource. We set the MLM probability to be 0.15.

\subsubsection{Sentiment Analysis Fine-Tuning}
Similar with baseline models, we also perform sentiment analysis task fine-tuning with the translated Financial Phrasebank dataset on GPU T4. We perform fine-tuning for sentiment analysis given 10\% of training data to 100\% of training data. This was performed to evaluate the domain-specific post-training effectiveness in transfer learning especially when there is a limited number of annotated dataset. The models are fine-tuned for 2 epochs with a learning rate of 2e-5 and a weight decay of 0.01.

Furthermore, we also perform another sentiment analysis fine-tuning for post-trained models by using IndoFinSent that is specific for one of the biggest financial institutions in Indonesia. The post-trained models, both base and large architecture, were fine-tuned with learning rate of 2e-5, batch size of 16, epoch number of 2, and weight decay of 0.01.

\subsubsection{Topic Classification Fine-Tuning}
In addition, we perform topic classification task fine-tuning with the translated dataset\footnote{https://huggingface.co/datasets/intanm/indonesian-financial-topic-classification-dataset}. Similar to the previous task, we also perfom fine-tuning given 10\% of training data to 100\% of training data. The models (base only) are fine-tuned for 2 epochs with a learning rate of 2e-5 and a weight decay of 0.01. We do not fine-tune the large models for this task due to limited hardware resource. 

\subsection{Evaluation Metrics}
We use F1 score to measure the performance of both baselines and post-trained models for sentiment analysis and topic classification tasks. The F1 score is the harmonic mean of the precision and recall, thus, it represents both precision and recall in a single metric. F1 score is used to provide a balanced evaluation across different classes, especially in situations where class imbalances exist.

\begin{table}
\centering
\caption{Sentiment Analysis Task F1 Score on Different Train Data Percentages: Base Arch.}
\label{table:5}
\begin{tabular}{p{.5cm} p{1cm} | p{1cm} p{1cm} p{1.5cm}}
\hline
\textbf{    } & \textbf{    } & \multicolumn{3}{c}{Post-Trained with} \\
\hline
\textbf{Train Data (\%)} & \textbf{Baseline (base-p1)} & \textbf{Financial News Articles} & \textbf{Corporate Financial Reports} & \textbf{Combination} \\
\hline
100 & 0.91 & \textbf{0.94$_{(+.03)}$} & \textbf{0.94$_{(+.03)}$} & \textbf{0.94$_{(+.03)}$} \\
90  & \textbf{0.92} & 0.91$_{(-.01)}$ & \textbf{0.92$_{(+0.0)}$} & 0.90$_{(-.02)}$  \\
80  & 0.87 & 0.92$_{(+.05)}$ & \textbf{0.93$_{(+.06)}$} & \textbf{0.93$_{(+.06)}$} \\
70  & 0.85 & \textbf{0.93$_{(+.08)}$} & 0.90$_{(+.05)}$ & 0.91$_{(+.06)}$ \\
60  & 0.82 & 0.90$_{(+.08)}$ & \textbf{0.92$_{(+.10)}$} & 0.89$_{(+.07)}$ \\
50  & 0.88 & \textbf{0.91$_{(+.03)}$} & 0.86$_{(-.02)}$ & 0.89$_{(+.01)}$ \\
40  & 0.79 & \textbf{0.89$_{(+.10)}$} & \textbf{0.89$_{(+.10)}$} & 0.87$_{(+.08)}$ \\
30  & 0.55 & \textbf{0.81$_{(+.26)}$} & 0.79$_{(+.24)}$ & 0.78$_{(+.23)}$ \\
20  & \textbf{0.63} & 0.55$_{(-.08)}$ & 0.55$_{(-.08)}$ & 0.57$_{(-.06)}$ \\
10  & 0.5  & \textbf{0.52$_{(+.02)}$} & 0.49$_{(-.01)}$ & 0.51$_{(+.01)}$ \\
\hline
\end{tabular}
\end{table}

\begin{table}
\centering
\caption{Sentiment Analysis Task F1 Score on Different Train Data Percentages: Large Arch.}
\label{table:6}
\begin{tabular}{p{1cm} p{1cm} | p{1cm} p{1cm} p{1.5cm}}
\hline
\textbf{    } & \textbf{    } & \multicolumn{3}{c}{Post-Trained with} \\
\hline
\textbf{Train Data (\%)} & \textbf{Baseline (large-p1)} & \textbf{Financial News Articles} & \textbf{Corporate Financial Reports} & \textbf{Combination} \\
\hline
100 & 0.96 & \textbf{0.97$_{(+.01)}$} & 0.95$_{(-.01)}$ & 0.95$_{(-.01)}$ \\
90  & 0.94 & \textbf{0.97$_{(+.03)}$} & 0.94$_{(+.00)}$ & 0.94$_{(+.00)}$ \\
80  & 0.95 & \textbf{0.96$_{(+.01)}$} & 0.94$_{(-.01)}$ & 0.94$_{(-.01)}$ \\
70  & 0.94 & \textbf{0.95$_{(+.01)}$} & 0.94$_{(+.00)}$ & 0.93$_{(-.01)}$ \\
60  & \textbf{0.94} & 0.92$_{(-.02)}$ & 0.93$_{(-.01)}$ & 0.90$_{(-.04)}$  \\
50  & \textbf{0.94} & \textbf{0.94$_{(+.00)}$} & 0.93$_{(-.01)}$ & 0.91$_{(-.03)}$ \\
40  & \textbf{0.89} & 0.85$_{(-.04)}$ & 0.86$_{(-.03)}$ & 0.86$_{(-.03)}$ \\
30  & 0.76 & \textbf{0.81$_{(+.05)}$} & 0.72$_{(-.04)}$ & 0.74$_{(-.02)}$ \\
20  & 0.52 & \textbf{0.68$_{(+.16)}$} & 0.58$_{(+.06)}$ & 0.55$_{(+.03)}$ \\
10  & 0.5  & 0.45$_{(-.05)}$ & \textbf{0.51$_{(+.01)}$} & 0.45$_{(-.05)}$ \\
\hline
\end{tabular}
\end{table}

\section{Results and Analysis}
In this section, we show the results of the post-trained IndoBERT for financial domain and analyze the performance of our models for sentiment analysis and topic classification downstream tasks. The code and post-trained models are available at https://github.com/intanq/indonesian-financial-domain-lm.

\subsection{Effectiveness of Domain-Specific Post-Training and Its Impact on Sentiment Analysis Downstream Task}
We can see in Table \ref{table:5}, that domain-specific post-trained IndoBERT (base architecture) outperforms the baseline IndoBERT base for most of the training data percentages on financial sentiment analysis task. It is also true for post-trained IndoBERT (large architecture) (see Table \ref{table:6}), however, baseline model (IndoBERT large) still outperforms the post-trained ones on some train data percentages such as 40\%, 50\%, and 60\%. Nevertheless, this shows the effectiveness of domain-specific post-training on pre-trained generic contextual language model for domain-specific downstream task and when given a smaller size of data for task-specific fine-tuning.

Interestingly, the post-trained base models have significant margins compared to the baseline models in terms of the F1 score, especially when using only 30\% of training data where it has a margin of 26\%. Meanwhile, the post-trained large models have a smaller margin compared to the baseline IndoBERT large. As we can see in Table \ref{table:6}, the performance of fine-tuning of the baseline large model itself is already high due to its larger number of parameters that affects its capacity to capture more linguistic features. From this analysis, we can see that the base model benefits more from the domain-specific post-training, enabling them to capture more relevant financial language patterns and nuances and therefore, increase the effectiveness on financial sentiment analysis downstream task by a bigger margin.

\subsection{Inference using IndoFinSent Dataset}
\begin{table}
\centering
\caption{F1 Score and Accuracy on IndoFinSent using Domain-Specific Post-Trained Models}
\label{table:7}
\begin{tabular}{p{3.5cm} | p{1cm} | p{1cm} | p{1cm}}
\hline
\textbf{Post-Training Dataset} & \textbf{LM Arch.} & \textbf{F1 Score} & \textbf{Accuracy}\\
\hline
Financial News Articles & Base & \textbf{0.81} & \textbf{0.81}\\
Corporate Financial Reports & Base & 0.80 & 0.81\\
Combined & Base & 0.79 & 0.80\\
Financial News Articles & Large & 0.79 & 0.80\\
Corporate Financial Reports & Large & 0.78 & 0.79\\
Combined & Large & 0.79 & 0.79\\
\hline
\end{tabular}
\end{table}
As mentioned in Section \ref{sec:fin-sa-dataset}, we also construct financial sentiment analysis dataset in Indonesian that is specific to one of the biggest financial institutions in Indonesia (IndoFinSent). The justification for creating a dataset in a target language rather than only use translated version of a dataset from a high-resource language (i.e., English) is to enable preservation of language nuances, cultural references, idiomatic expressions, and wordplay that are specific to Indonesian language. Moreover, by not only relying on translated dataset, it enables us to avoid translation errors or loss of meaning that can lead to incorrect or misleading results. We use this data to perform another fine-tuning for sentiment anaysis task on our domain-specific post-trained models.

From Table \ref{table:7}, we can see that fine-tuning post-trained IndoBERT using our constructed data result in similar performance. This is an interesting finding since the fine-tuning performance of large models as shown in Table \ref{table:6} that uses translated Financial Phrasebank data outperforms the fine-tuned post-trained base models in general. 

\subsection{Effectiveness of Domain-Specific Post-Training on Topic Classification Downstream Task}
\begin{table}
\centering
\caption{Topic Classification Task F1 Score on Different Train Data Percentages: Base Arch.}
\label{table:topic-clf-exp}
\begin{tabular}{p{.5cm} p{1cm} | p{1cm} p{1cm} p{1.5cm}}
\hline
\textbf{    } & \textbf{    } & \multicolumn{3}{c}{Post-Trained with} \\
\hline
\textbf{Train Data (\%)} & \textbf{Baseline (base-p1)} & \textbf{Financial News Articles} & \textbf{Corporate Financial Reports} & \textbf{Combination} \\
\hline
100 & \textbf{0.85} & \textbf{0.85$_{(+.00)}$} & \textbf{0.85$_{(+.00)}$} & \textbf{0.85$_{(+.00)}$}\\
90 & \textbf{0.84} & 0.8$_{(-.04)}$ & 0.8$_{(-.04)}$ & \textbf{0.84$_{(+.00)}$}\\
80 & 0.81 & \textbf{0.83$_{(+.02)}$} & \textbf{0.83$_{(+.02)}$} & 0.82$_{(+.01)}$\\
70 & 0.76 & \textbf{0.81$_{(+.05)}$} & 0.76$_{(+.00)}$ & 0.78$_{(+.02)}$\\
60 & 0.77 & \textbf{0.79$_{(+.02)}$} & \textbf{0.79$_{(+.02)}$} & \textbf{0.79$_{(+.02)}$}\\
50 & 0.74 & \textbf{0.78$_{(+.04)}$} & 0.72$_{(-.02)}$ & 0.77$_{(+.03)}$\\
40 & 0.69 & 0.69$_{(+.00)}$ & 0.67$_{(-.02)}$ & \textbf{0.7$_{(+.01)}$}\\
30 & 0.64 & \textbf{0.66$_{(+.02)}$} & 0.63$_{(-.01)}$ & 0.64$_{(+.00)}$\\
20 & 0.56 & \textbf{0.58$_{(+.02)}$} & 0.53$_{(-.03)}$ & 0.57$_{(+.01)}$\\
10 & 0.33 & \textbf{0.37$_{(+.04)}$} & 0.3$_{(-.03)}$ & 0.28$_{(-.05)}$\\

\hline
\end{tabular}
\end{table}
As we can see in Table \ref{table:topic-clf-exp}, most of the models post-trained by financial news articles dataset outperform the IndoBERT baseline. It can also be observed that in lower training data percentages, the post-trained models outperform the baseline. This shows the effectiveness of domain-specific post-training in topic classification task when given a smaller size of annotated data for fine-tuning. However, it does not apply to the models post-trained by corporate financial reports and combination (i.e., financial news articles and corporate financial reports) where the baselines still outperform most of the models. Although the margin between the post-trained models and the baselines are quite close for the outperforming ones (i.e., ranging from 0\% to 5\%), it still shows that domain-specific post-training does impact the effectiveness of contextual language model in domain-specific downstream tasks.

\subsection{Post-Training Corpus: Financial News Articles vs Financial Corporate Reports}
There are two types of financial texts that we use for domain-specific post-training. The language styles of the texts are mixed and formal for financial news articles and corporate financial reports respectively. This is done to enforce the diversity of the corpus in terms of language styles.

Furthermore, it can be noticed that the post-trained models using financial news articles data outperform other models in general. This is caused by the language style similarity between the post-training corpus and downstream tasks dataset (i.e., sentiment analysis and topic classification) that are mainly from news and social media domain. Thus, the results might be different if the data used for sentiment analysis or topic classification fine-tuning has a different language style or gathered from other than news domain.

\section{Conclusion}
We perform contextual language model post-training for financial domain by using masked language model training objective. Financial news articles and corporate financial reports are used as the unlabeled corpus for the post-training. In our experiment scenario, we post-train the IndoBERT model (both base and large architecture) by using either financial news articles, corporate financial reports, and combination of both. The post-trained models were evaluated by fine-tuning them to financial sentiment analysis and topic classification downstream tasks using the annotated dataset with 10 different training sizes. This aims to show the effectiveness of the domain-specific post-training especially when given a limited number of annotated dataset.

The experiments reveal interesting findings. Overall, the post-training of IndoBERT models for financial domain helps to improve the performance of the mentioned downstream tasks. For sentiment analysis task, the IndoBERT base model benefits more from this domain-specific post-training. However, it is observed that the domain-specific post-training for IndoBERT large model only improve its performance on downstream task by a small margin. This is due to its initial capacity and a larger number of parameters already allows the large model to capture more linguistic features. In addition, topic classification fine-tuning of the post-trained models also shows the effectiveness of the post-training.

\section{Future Works}
For future works, from-scratch pre-training can be performed for various language models and thus, a larger size of unlabeled text data in financial domain (Indonesian) will be required. This is also need to be supported by sufficient GPU resources. Moreover, regarding the construction of financial sentiment analysis in Indonesian, it can also be gathered from other than news domain, such as, online forums, financial reports, customer inquiries, and more.

\section*{Acknowledgment}
We would like to thank Vina Alvionita for participating in the construction of the IndoFinSent sentiment analysis dataset. We also would like to thank our other colleagues from Digital Banking Development \& Operation Division, PT Bank Rakyat Indonesia (Persero) Tbk for the constructive feedback to enhance this work.


\bibliographystyle{unsrt}
\bibliography{custom}

\begin{thebibliography}{10}

\bibitem{wilie-etal-2020-indonlu}
Bryan Wilie, Karissa Vincentio, Genta~Indra Winata, Samuel Cahyawijaya,
  Xiaohong Li, Zhi~Yuan Lim, Sidik Soleman, Rahmad Mahendra, Pascale Fung,
  Syafri Bahar, and Ayu Purwarianti.
\newblock {I}ndo{NLU}: Benchmark and resources for evaluating {I}ndonesian
  natural language understanding.
\newblock In {\em Proceedings of the 1st Conference of the Asia-Pacific Chapter
  of the Association for Computational Linguistics and the 10th International
  Joint Conference on Natural Language Processing}, pages 843--857, Suzhou,
  China, December 2020. Association for Computational Linguistics.

\bibitem{devlin-etal-2019-bert}
Jacob Devlin, Ming-Wei Chang, Kenton Lee, and Kristina Toutanova.
\newblock {BERT}: Pre-training of deep bidirectional transformers for language
  understanding.
\newblock In {\em Proceedings of the 2019 Conference of the North {A}merican
  Chapter of the Association for Computational Linguistics: Human Language
  Technologies, Volume 1 (Long and Short Papers)}, pages 4171--4186,
  Minneapolis, Minnesota, June 2019. Association for Computational Linguistics.

\bibitem{huang-finbert}
Allen~H. Huang, Hui Wang, and Yi~Yang.
\newblock Finbert: A large language model for extracting information from
  financial text*.
\newblock {\em Contemporary Accounting Research}, 40(2):806--841, 2023.

\bibitem{yang2020finbert}
Yi~Yang, Mark Christopher~Siy UY, and Allen Huang.
\newblock Finbert: A pretrained language model for financial communications,
  2020.

\bibitem{araci2019finbert}
Dogu Araci.
\newblock Finbert: Financial sentiment analysis with pre-trained language
  models, 2019.

\bibitem{malo2013good}
Pekka Malo, Ankur Sinha, Pyry Takala, Pekka Korhonen, and Jyrki Wallenius.
\newblock Good debt or bad debt: Detecting semantic orientations in economic
  texts, 2013.

\bibitem{peters-etal-2018-deep}
Matthew~E. Peters, Mark Neumann, Mohit Iyyer, Matt Gardner, Christopher Clark,
  Kenton Lee, and Luke Zettlemoyer.
\newblock Deep contextualized word representations.
\newblock In {\em Proceedings of the 2018 Conference of the North {A}merican
  Chapter of the Association for Computational Linguistics: Human Language
  Technologies, Volume 1 (Long Papers)}, pages 2227--2237, New Orleans,
  Louisiana, jun 2018. Association for Computational Linguistics.

\bibitem{howard2018universal}
Jeremy Howard and Sebastian Ruder.
\newblock Universal language model fine-tuning for text classification, 2018.

\bibitem{yang2020xlnet}
Zhilin Yang, Zihang Dai, Yiming Yang, Jaime Carbonell, Ruslan Salakhutdinov,
  and Quoc~V. Le.
\newblock Xlnet: Generalized autoregressive pretraining for language
  understanding, 2020.

\bibitem{radford2018improving}
Alec Radford, Karthik Narasimhan, Tim Salimans, Ilya Sutskever, et~al.
\newblock Improving language understanding by generative pre-training.
\newblock 2018.

\bibitem{gu-biomedical}
Yu~Gu, Robert Tinn, Hao Cheng, Michael Lucas, Naoto Usuyama, Xiaodong Liu,
  Tristan Naumann, Jianfeng Gao, and Hoifung Poon.
\newblock Domain-specific language model pretraining for biomedical natural
  language processing.
\newblock {\em ACM Trans. Comput. Healthcare}, 3(1), oct 2021.

\bibitem{Lee_2019}
Jinhyuk Lee, Wonjin Yoon, Sungdong Kim, Donghyeon Kim, Sunkyu Kim, Chan~Ho So,
  and Jaewoo Kang.
\newblock {BioBERT}: a pre-trained biomedical language representation model for
  biomedical text mining.
\newblock {\em Bioinformatics}, 36(4):1234--1240, sep 2019.

\bibitem{peng-etal-2019-transfer}
Yifan Peng, Shankai Yan, and Zhiyong Lu.
\newblock Transfer learning in biomedical natural language processing: An
  evaluation of {BERT} and {ELM}o on ten benchmarking datasets.
\newblock In {\em Proceedings of the 18th BioNLP Workshop and Shared Task},
  pages 58--65, Florence, Italy, August 2019. Association for Computational
  Linguistics.

\bibitem{beltagy-etal-2019-scibert}
Iz~Beltagy, Kyle Lo, and Arman Cohan.
\newblock {S}ci{BERT}: A pretrained language model for scientific text.
\newblock In {\em Proceedings of the 2019 Conference on Empirical Methods in
  Natural Language Processing and the 9th International Joint Conference on
  Natural Language Processing (EMNLP-IJCNLP)}, pages 3615--3620, Hong Kong,
  China, November 2019. Association for Computational Linguistics.

\bibitem{chalkidis-etal-2020-legal}
Ilias Chalkidis, Manos Fergadiotis, Prodromos Malakasiotis, Nikolaos Aletras,
  and Ion Androutsopoulos.
\newblock {LEGAL}-{BERT}: The muppets straight out of law school.
\newblock In {\em Findings of the Association for Computational Linguistics:
  EMNLP 2020}, pages 2898--2904, Online, November 2020. Association for
  Computational Linguistics.

\end{thebibliography}

\end{document}